\documentclass[conference]{IEEEtran}
\IEEEoverridecommandlockouts
\usepackage{cite}
\usepackage{amsmath,amssymb,amsfonts}
\usepackage{algorithmic}
\usepackage{graphicx}
\usepackage{textcomp}
\usepackage{xcolor}
\def\BibTeX{{\rm B\kern-.05em{\sc i\kern-.025em b}\kern-.08em
    T\kern-.1667em\lower.7ex\hbox{E}\kern-.125emX}}

\newcommand{\model}{SCM-GAN}
\newcommand{\vcvoice}{CycleGAN-VC$^{\text{sp}}$}
\newcommand{\vcsong}{CycleGAN-VC$^{\text{voc}}$}
\newcommand{\vcscratch}{CycleGAN-VC$^{\text{voc}}_{\text{scratch}}$}
\newcommand{\vcnosplit}{CycleGAN-VC$^{\text{voc+music}}$}    
    
\begin{document}

\title{Change Your Singer: A Transfer Learning Generative Adversarial Framework for Song to Song Conversion \\
\thanks{*Authors with equal contribution}
}

\author{\IEEEauthorblockN{Rema Daher*}
\IEEEauthorblockA{\textit{Dept. of Mechanical Engineering} \\
\textit{American University of Beirut}\\
Beirut, Lebanon \\
rgd05@mail.aub.edu}
\and
\IEEEauthorblockN{Mohammad Kassem Zein*}
\IEEEauthorblockA{\textit{Dept. of Mechanical Engineering} \\
\textit{American University of Beirut}\\
Beirut, Lebanon \\
mhk50@mail.aub.edu}
\and
\IEEEauthorblockN{Julia El Zini}
\IEEEauthorblockA{\textit{Dept. of Electrical and Computer Engineering} \\
\textit{American University of Beirut}\\
Beirut, Lebanon \\
jwe04@mail.aub.edu}
\and
\IEEEauthorblockN{Mariette Awad}
\IEEEauthorblockA{\textit{Dept. of Electrical and Computer Engineering} \\
\textit{American University of Beirut}\\
Beirut, Lebanon \\
ma162@mail.aub.edu}
\and
\IEEEauthorblockN{Daniel Asmar}
\IEEEauthorblockA{\textit{Dept. of Mechanical Engineering} \\
\textit{American University of Beirut}\\
Beirut, Lebanon \\
da20@mail.aub.edu}
}

\maketitle

\begin{abstract}
\textit{Have you ever wondered how a song might sound if performed by a different artist?} In this work, we propose \model, an end-to-end non-parallel song conversion system powered by generative adversarial and transfer learning, which allows users to listen to a selected target singer singing \textit{any} song. \model~first separates songs into vocals and instrumental music using a U-Net network, then converts the vocal segments to the target singer using advanced CycleGAN-VC, before merging the converted vocals with their corresponding background music. \model~is first initialized with feature representations learned from a state-of-the-art voice-to-voice conversion and then trained on a dataset of non-parallel songs. After that, \model~is evaluated against a set of metrics including global variance GV and modulation spectra MS on the 24 Mel-cepstral coefficients (MCEPs). Transfer learning improves the GV by 35\% and the MS by 13\% on average. A subjective comparison is conducted to test the output's similarity to the target singer and its naturalness. Results show that the \model's similarity between its output and the target reaches 69\%, and its naturalness reaches 54\%.% and the positive effect of splitting vocals from background music on the performance of \model. %e evaluation metrics, we prove the importance of each section of the pipeline to the contribution of the whole system.

%metrics. The objective metrics included an analysis of the global variance and modulation spectra on the 24 Mel-cepstrum coefficients of the results. These metrics along with the subjective results showed that the output is natural and and very similar to the ground truth.
\end{abstract}


\section{Introduction}\label{sec:intro}
%Text To Speech (TTS) systems have gained much interest in the last decades, especially with the increased %interest in audiobooks, e-learning systems, and speaking robots. Voices created by the TTS scheme are not %usually human-like, but with the progress in deep networks, a given text can now be spoken by any selected %human voice \cite{ping2017deep}.
Voice-to-voice conversion is the process of converting a speech spoken by a particular speaker to another selected target speaker. Prior work on voice-to-voice using deep learning utilized sequence-to-sequence voice conversion \cite{kaneko2017sequence}, and phoneme-based linear mapping functions \cite{liu2007high} to provide an end-to-end voice-to-voice solution. Recently, generative adversarial networks (GANs) have shown their success in natural language processing \cite{hu2017toward}, and image and video synthesis \cite{xu2018attngan}.
Given the requirement of generating a voice that mimics a particular data distribution, systems based on GANs  \cite{binkowski2019high} showed promising results in voice-to-voice conversions.

Song-to-song systems are a particular case of the voice-to-voice problem, and attempt to change existing songs by incorporating the voice of a user-selected artist. Such systems have many practical applications. For instance, music applications can integrate novel features that allow users to listen to any song by the voice of their favorite singer. Additionally, users can pretend on social media platforms to sing a song by replacing the voice of the original singer by their own voice. 

Given that speech and music encode distinct sorts of information differently, their acoustical features are fundamentally dissimilar \cite{wolfe2002speech}. For instance, fundamental frequencies, temporal regularities and quantization, short silences, steady and varying formants, and transient spectral details significantly vary between speech and music. This makes speech recognition challenging when there is even a modest level of background music \cite{wolfe2002speech}. Existing song-to-song systems only focus on achieving the singing voice conversion without developing stand-alone end-to-end systems that perform well when background music is inputted along the vocals \cite{kobayashi2015statistical}. 

In this work, we propose a novel end-to-end system powered by generative adversarial networks and transfer learning, and which replaces the original singer of a song by any desired performer. The long term objective of the proposed system aims at enabling developers to build a commercial application for the aforementioned purpose. Our model, \textbf{S}plit-\textbf{C}onvert-\textbf{M}erge using Cycle\textbf{GAN}, \model, first takes advantage of the U-Net \cite{jansson2017singing} to split vocals from the background music, then trains an instance of Voice Converter CycleGANs \cite{kaneko2018parallel} on a set of in-house collected songs, not necessarily parallel. Finally, \model~merges the converted singing voice with the background music to achieve the song-to-song conversion. Moreover, we utilize acoustic features learnt in the voice converter CycleGAN of \cite{kaneko2018parallel} to efficiently train \model~using transfer learning. We show the importance of the suggested system through objective and subjective evaluation. The results show that \model~successfully converts a song to a target singer song with high resemblance to ground truth. Transfer learning improves the average GV and MS by 35\% and 13\% respectively and the splitting scheme increases the subjective evaluation scores by 73\% and 26\% on the naturalness and similarity of the converted song. The contributions of this work include: (1) an end to end song-to-song conversion approach which (2) combines two mostly-unrelated machine learning tasks with co-dependent results. The purpose of this combination is to (3) split and reconstruct songs as well as (4) transfer knowledge from voice-to-voice models.

The remainder of the paper is organized as follows: first, related work is reported in Section \ref{sec:lit_rev}, then Section \ref{sec:method} describes the system Methodology before Section \ref{sec:results} reports the results of the conducted experiments. Section \ref{sec:conc} concludes with final remarks. 
\section{Related Work}
\label{sec:lit_rev}
Converting a specific speaker's voice to a target voice is a topic of interest for many researchers. For instance, \cite{ohtani2006maximum} presented a voice conversion technique by introducing a STRAIGHT mixed excitation \cite{kawahara1999restructuring} to Maximum Likelihood Estimation (MLE) with a Gaussian Mixture model (GMM). However, in their work they focused on the quality of the converted voice rather than the conversion accuracy. Another approach of voice conversion was introduced by \cite{desai2009voice}. Their system involved mapping the spectral features of a source speaker to that of a target speaker using Artificial Neural Network (ANN); they proved that the mapping capabilities of (ANN) perform better transformation and produce better quality voices than GMMs. 

Deep Neural Networks (DNNs) were also used in voice conversion systems \cite{nakashika2013voice}. However, the problem with conventional DNN frame-based methods is that they do not capture the temporal dependencies of a speech sequence. To tackle this problem, \cite{sun2015voice} proposed a Deep Bidirectional Long Short-Term Memory based Recurrent Neural Network (DBLSTM-RNN) architecture to model the long-range context-dependencies in the acoustic trajectory between the source and the target voice. Recently, Kaneko and Kameoka in \cite{kaneko2018parallel} proposed a novel voice converter (CycleGAN-VC) that relies on cycle-consistent adversarial networks (CycleGAN). 

%Converting a specific speaker's voice to a target voice has been a topic of interest for many researchers. For instance, \cite{ohtani2006maximum} used a Gaussian Mixture model (GMM) to make the conversion; later, \cite{desai2009voice} relied on an Artificial Neural Network (ANN) to achieve better results; after that, \cite{nakashika2013voice} used a Deep Neural Networks (DNNs) for the conversion and outperformed the previous two systems. All of these previously mentioned methods do not capture the temporal dependencies of a speech sequence, and to tackle this problem, \cite{sun2015voice} proposed a Deep Bidirectional Long Short-Term Memory based Recurrent Neural Network (DBLSTM-RNN). Recently, \cite{kaneko2018parallel} proposed a novel voice converter (CycleGAN-VC) that relies on cycle-consistent adversarial networks (CycleGAN). 

%During the last few years, researchers started studying the application of voice conversion to
For the conversion of voices that are sung, \cite{villavicencio2010applying} relied on statistical modeling of the speakers' timbre space using a GMM in order to define a time-continuous mapping from the feaures of the the source speaker to the target. Moreover, \cite{kobayashi2014statistical} used direct waveform modification based on the spectrum differential to achieve the conversion of the singing voice. Then, they extended it to restore the global variance of the converted spectral parameter trajectory to avoid over-smoothing at unvoiced frames in \cite{kobayashi2015statistical}.

So far, the attempts of conversion of singing voice have relied on methods used for voice-to-voice conversion. However, to the best of our knowledge, the state of the art method in voice conversion, CycleGAN-VC \cite{kaneko2018parallel}, has not yet been implemented for the conversion of singing voice. In this paper, we propose an end-to-end system \model~that employs CycleGAN-VC \cite{kaneko2018parallel} along with a deep U-Net \cite{jansson2017singing} to achieve song-to-song conversion.

\section{Methodology}\label{sec:method}
%\begin{figure}[h]
%\centering
%\includegraphics[width=0.49\textwidth]{figs/training_procedure.png}
%\caption{}
%\label{fig:training}
%\end{figure}
In this work, we propose \model, an end-to-end system that converts songs from the voice of \textit{any} singer to that of a \textit{specific fixed} target singer $\mathcal{S}$ without altering the background music. For this purpose, \vcsong, a deep CycleGAN converter that is trained on instances of the form ($voc_{i,A}, voc_{i,\mathcal{S}}$) where $voc_{i,S}$ is the $i^{\text{th}}$ vocals segment sung by the singer $S$. % and $\mathcal{S}$ is the \textit{fixed} target singer. 
To be able to maintain the background music, only vocals are fed into \vcsong~after being separated from the background music by a deep U-Net \cite{jansson2017singing}. 

\begin{figure*}
    \centering
    \includegraphics[width=\textwidth]{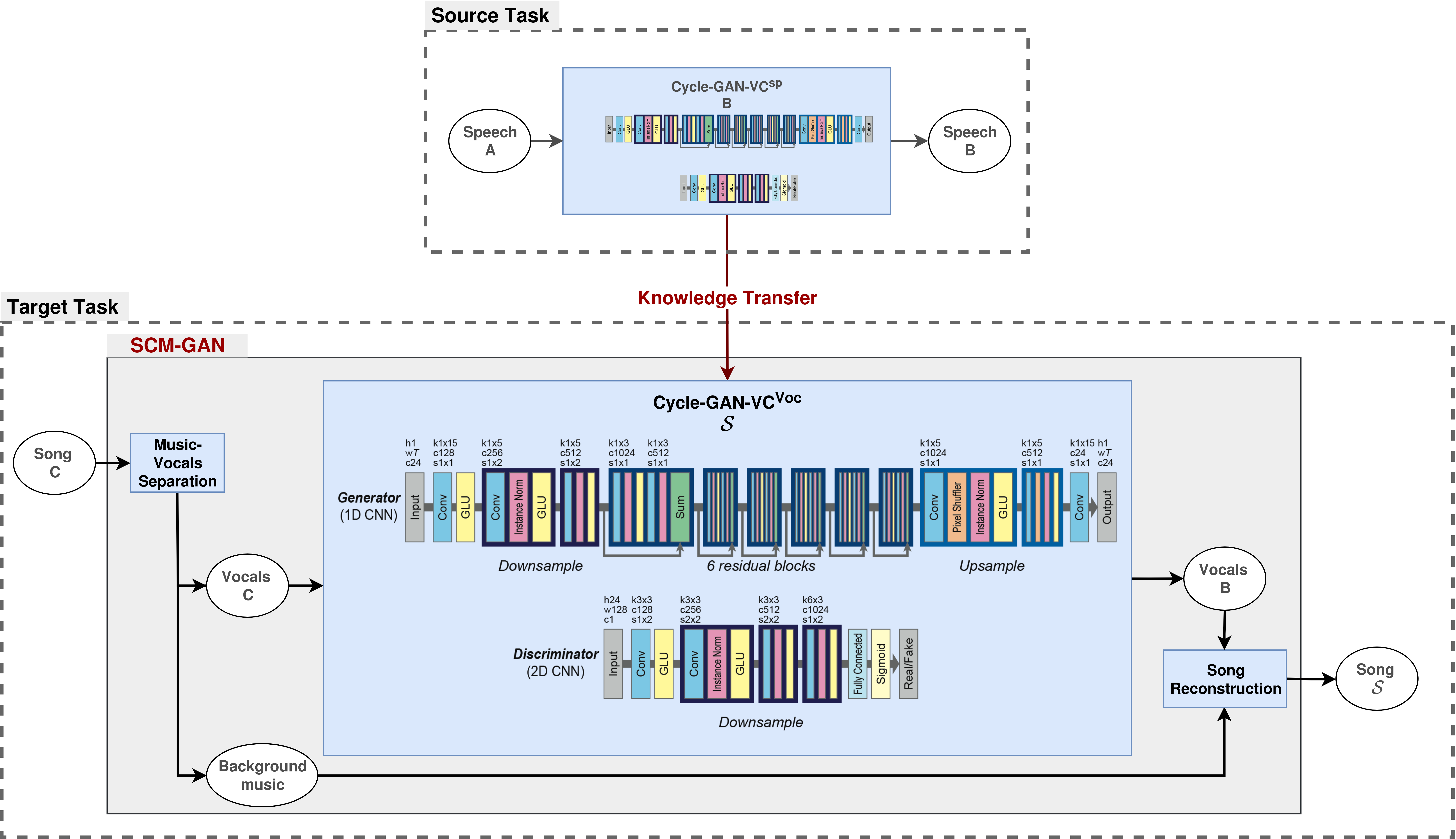}
    \caption{SCM-GAN system overview. \cite{kaneko2018parallel} provides details of the adopted cycle-GAN architecture}
    \label{fig:system}
\end{figure*}

The overall workflow of \model~is shown in Fig.~\ref{fig:system}. First, the song is fed into a pre-trained U-Net model \cite{jansson2017singing}, which separates songs into vocals and background music. Then, the vocals are inputted into the \vcsong, which converts them into the voice of $\mathcal{S}$ using transfer learning. Finally, a merging scheme is used to overlay the converted output with the saved background music from the separation phase.
%-----------
%------------
The system is composed of four main components including: (1) the vocals-music separation with a pre-trained U-Net \cite{jansson2017singing} (2) the vocals conversion provided by CycleGAN-VC, (3) the knowledge transferred from a CycleGAN-VC network trained on \textit{speech} and (4) the merging scheme. 
Four different notations will be used throughout the paper to distinguish between CycleGAN-VC implementations: (1) \vcvoice~trained on speech, 2) \vcsong~trained on vocals using transfer learning, 3) \vcscratch~trained on vocals from scratch and 4) \vcnosplit~trained on song (vocals and background music) using transfer learning.\\

\subsection{Music-Vocals Separation}\label{sec:method_split}
A U-Net \cite{jansson2017singing} is implemented as an encoder-decoder fully connected convolutional neural network to separate the background music from the vocals by operating exclusively on the magnitude of audio spectrograms. Specifically, the U-Net implements two decoders: one for the instrumental music and another one for the vocals. The audio signal for both components (instrumental/vocal) is recompiled as follows: the magnitude component of the signal is reconstructed by applying the output mask of each decoder to the magnitude of the original spectrum; and its phase component is that of the original spectrum without any modifications. 
% \begin{equation}
% L(X,Y;\ominus) = \parallel f(X,\ominus) \odot X - Y \parallel_{1,1}
% \end{equation}

% with $X$ as the magnitude of the original song spectogram and $Y$ as that of the target output. $f(X,\ominus )$ is the model output with the model generated mask parameter $\ominus$.

% In fact, the pretrained model is obtained by training 2 U-Nets to estimate the vocal along with the instrumental mask.

\subsection{Vocals Conversion With \vcsong}\label{sec:method_conv}
In this work, vocals are converted by \vcsong~(trained on vocals) that has the same underlying architecture as  in \cite{kaneko2018parallel}. The CycleGAN-VC architecture modifies that of cycleGAN \cite{zhu2017unpaired}, adding to it identity mapping loss \cite{taigman_2017_ICLR} and a gated CNN \cite{dauphin2017language} that can represent sequential and hierarchical features of speech, and generate state-of-the-art speech output \cite{kaneko2017sequence}. The superior results are achieved because of the networks' structure, which include gated linear units (GLUs) that act as data driven activation functions:
\begin{equation*}
    H_{l+1}=(H_l*W_l+b_l) \otimes \sigma (H_l*V_l+c_l), 
\end{equation*}
where $H_{l+1}$ and $H_l$ are the $l+1$ and $l$ layer outputs respectively. In addition, $W_l$, $b_l$, $V_l$, and $c_l$ represent the parameters of the model and $\sigma$ represents the \textit{sigmoid} function. Here, $\otimes$ is the element-wise product.

In the CycleGAN-VC architecture, three losses are utilized, including an adversarial loss, a cycle-consistency loss, and an identity-mapping loss \cite{taigman_2017_ICLR}. First, we denote the mapping from the source $x \in X$ to the target $y \in Y$ as $G_{X \to Y}$ and the reciprocal as $G_{Y \to X}$. Then, the adversarial loss can be described as the difference between the distribution of converted data $P_{G_{X \to Y}}(x)$ and their corresponding actual training output distribution $P_{Data}(y)$. In order to reduce this difference and deceive the discriminator $D_Y$ by getting an output close to the target output, the following objective function is minimized:
%--------------------
%\begin{figure}
%\centering
%\includegraphics[width=0.49\textwidth]{figs/teaser_high_res.png}
%\caption{CycleGAN-VC Architecture Adopted From %\cite{kaneko2018parallel}}
%\label{fig:arch}
%\end{figure}
%--------------------
\begin{equation*}
\begin{split}
L_{adv}(G_{X \to Y},D_Y) = E_{y \sim P_{Data}(y)}[logD_Y(y)] + \\
E_{x \sim P_{Data}(x)}[log(1-D_Y(G_{X \to Y}(x)))]
\end{split}
\end{equation*}

Conversely, $D_Y$ maximizes this loss to avoid being deceived. Cycle-consistency loss attempts to keep the contextual information between the input and the converted output consistent using the following objective function:
\begin{equation*}
\begin{split}
L_{cyc}(G_{X \to Y},G_{Y \to X}) =\\E_{x \sim P_{Data}(x)}[\parallel G_{Y \to X}(G_{X \to Y}(x))-x \parallel_1]+
\\E_{y \sim P_{Data}(y)}[\parallel G_{X \to Y}(G_{Y \to X}(y))-y \parallel_1]
\end{split}
\end{equation*}
Identity-mapping loss is used to preserve composition and linguistic information between input and output. This loss is defined as follows:
%----------------
\begin{equation*}
\begin{split}
L_{id}(G_{X \to Y},G_{Y \to X}) = E_{y \sim P_{Data}(y)}[\parallel G_{X \to Y}(y)-y \parallel_1] +
\\E_{x \sim P_{Data}(x)}[\parallel G_{Y \to X}(x)-x \parallel_1]
\end{split}
\end{equation*}
%----------------
Using an inverse adversarial loss $L_{adv}(G_{Y \to X},D_X)$, the full objective function can be expressed as:
%----------------
\begin{equation*}
\begin{split}
L_{full} = L_{adv}(G_{X \to Y},D_Y) + L_{adv}(G_{Y \to X},D_X) +
\\ \lambda _{cyc}L_{cyc}(G_{X \to Y},G_{Y \to X})  + \lambda _{id}L_{id}(G_{X \to Y},G_{Y \to X})
\end{split}
\end{equation*}
%------------------
\noindent where $\lambda _{cyc}$ and $\lambda _{id}$ are trade-off parameters for their corresponding losses. This objective function allows the model to learn the mapping from a source singer to a target one chosen by the user.
%------------------

\subsection{Knowledge Transfer from \vcvoice}
\vcvoice~\cite{kaneko2018parallel} is trained on instances of the form $(\text{sp}_{i,A}, \text{sp}_{i,B})$ where $\text{sp}_{i,S}$ is the $i^{\text{th}}$ speech training instance spoken by $S$. In order to speed up the training of \vcsong, voice feature representation learnt in \vcvoice~is used to initialize the training. It is worth mentioning that the target speaker in \vcvoice~is different than $\mathcal{S}$, the target speaker of our \vcsong.  
%---------------
\subsection{Song Reconstruction}
Since the proposed pipeline maintains the temporal characteristic of the input audio after the splitting and converting step, it is enough to just overlay the background music with the output from the model. In order to overlay the converted vocals with their corresponding instrumental music, they are both segmented via an analysis of signal onsets and offsets as in \cite{hu2007auditory}. Segments are then integrated at a coarse scale and at a finer scale by locating accurate onset and offset positions for segments as in \cite{pydub}.
%------------ 
\section{Experiments}\label{sec:results}
To assess the efficiency of \model, we performed several experiments in which a performer is replaced by another singing the identical song. Since singing voice conversion methods have relied on voice-to-voice conversion, we will be comparing \model~to the state of the art method in voice-to-voice conversion (CycleGAN-VC). Given that no public dataset exists that includes two voices singing the same part with background music, we developed our own. 

\subsection{Dataset}
We collected a dataset of non-parallel aligned song segments of $3sec$ each on average to be consistent with the Voice Conversion Challenge 2016 (VCC 2016) dataset \cite{toda2016voice} used by \cite{kaneko2018parallel}. For every instance of a singer $A$, a corresponding instance is created with the target singer $\mathcal{S}$ singing the same lyrics at a different time frame. Particularly, $228$ training instances were created with Samantha Harvey as singer $A$ and Ed Sheeran as target singer $\mathcal{S}$. As for the testing data, $15$ instances of $10secs$ each (for better subjective and objective evaluation) were collected. These instances include songs by singer $A$ singing her own songs as opposed to singing singer $B$'s songs. In addition, songs from $5$ singers different from $A$ were also included in the testing data including: Beyonce, Bea Miller, Diamond White, Nicole Cross, and Chelsea FreeCoustic. This dataset will be made publicly available for further research and improvements in the field. 

After splitting the data into vocals and background music, the vocals were then pre-processed by downsampling the data to 16 kHz. Afterwards, at every 5 ms the data is transformed into MCEPs, aperiodicities (APs), and logarithmic fundamental frequency $log(F_0)$, using a speech synthesis system WORLD \cite{morise2016world}. Then, a normalized logarithm Gaussian transformation, \cite{liu2007high}, is applied on $F_0$ as in \cite{kaneko2018parallel}.

\subsection{Training}
After the separated vocals are preprocessed, voice-to-voice inter-gender weights ($SF1-TF2$) from \cite{kaneko2018parallel} are loaded into CycleGAN-VC, which is then fine-tuned on 1000 epochs using our data. The choice of the number of epochs was chosen to be 1 since the losses converged after that. Fine-tuning can be used in this case since the task the model has already learnt (voice-to-voice) is similar to the new task it is about to learn (song-to-song). The reason behind fine-tuning is that much less epochs and processing time are needed than that needed to train a model from scratch.

\subsection{Objective Evaluation}
\begin{figure}[bht]
\begin{center}
\begin{tabular}{c}
\includegraphics[width=0.65\columnwidth]{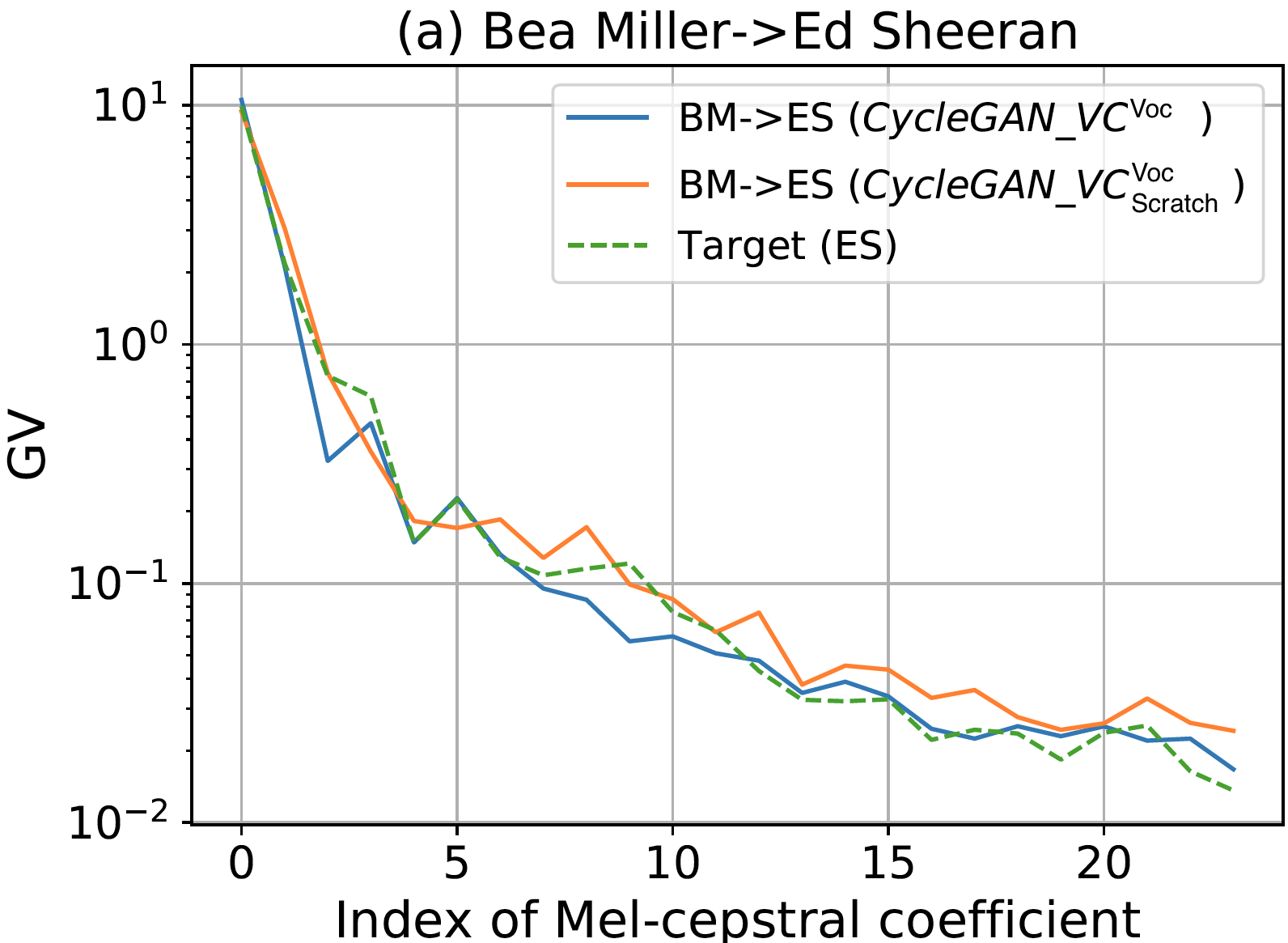}\\
\includegraphics[trim={0.18cm 0 0 0}, clip, width=0.65\columnwidth]{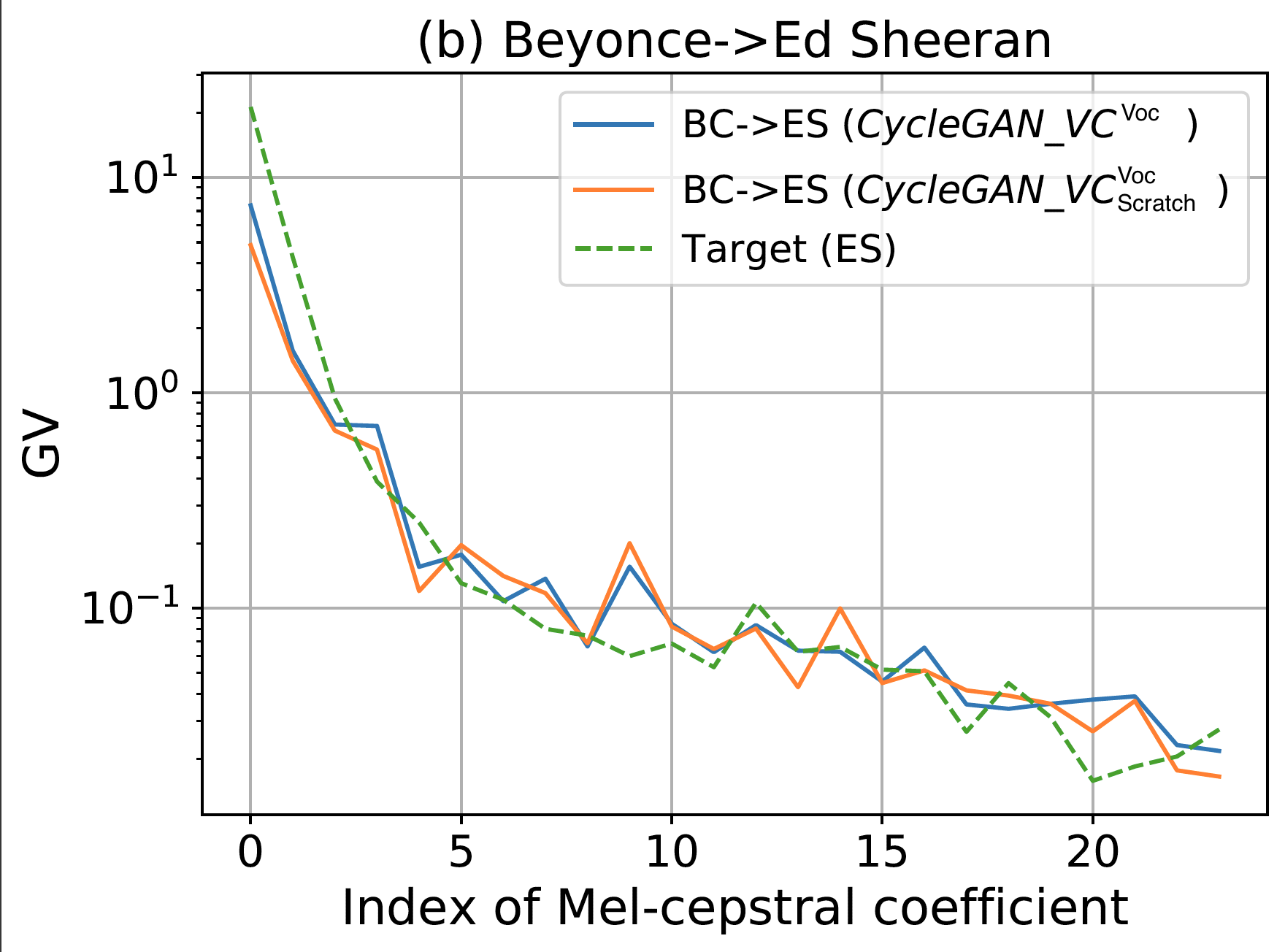}\cr
\end{tabular}
\end{center}
%\vspace{-0.1cm}
\caption{GV of \vcsong~and \vcscratch~on two song segments (best seen in color)}
\label{fig:GV1}
\end{figure}

\begin{figure}[htb]
\begin{center}
\begin{tabular}{c}
\includegraphics[width=0.65\columnwidth]{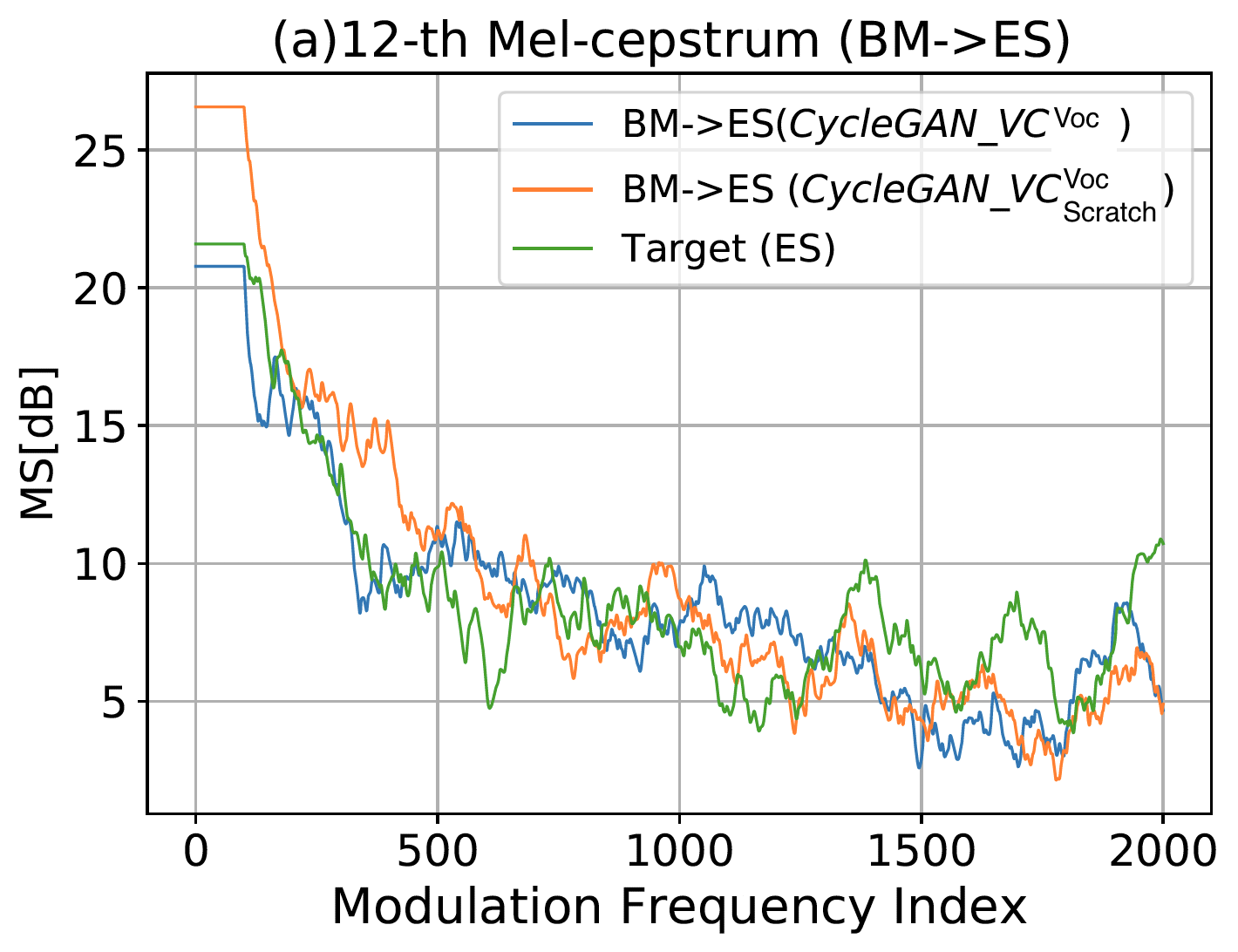}\\
\includegraphics[width=0.65\columnwidth]{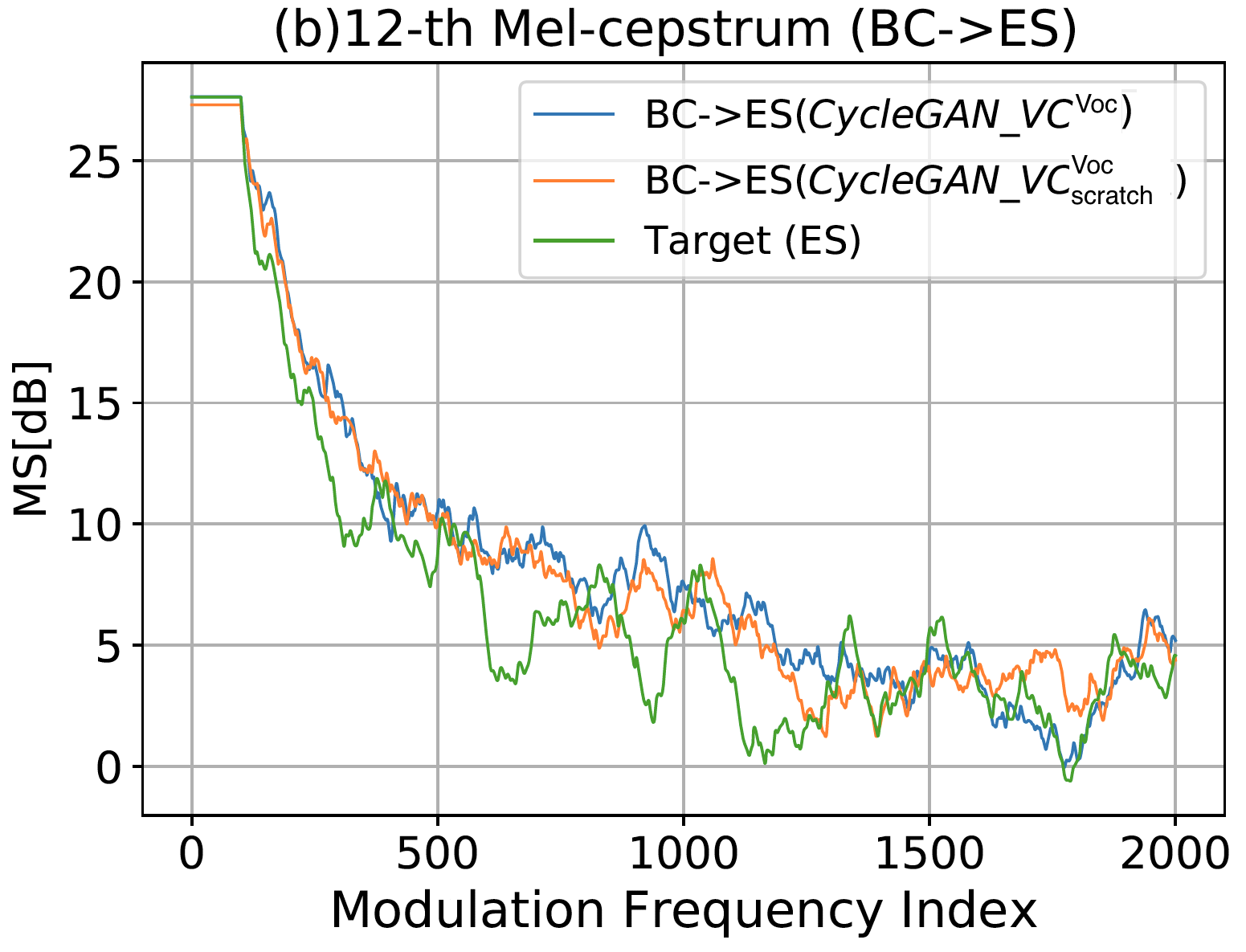}\cr
\end{tabular}
\end{center}
%\vspace{-0.1cm}
\caption{MS of \vcsong~and \vcscratch~on two song segments (best seen in color)}
\label{fig:MS1}
\end{figure}

To properly assess our proposed system, we evaluated the quality of the converted feature vector (MCEPs). Specifically, we focused in our experiments on analyzing two associated metrics: Modulation Spectrum (MS) \cite{takamichi2014postfilter} and Global Variance (GV) \cite{toda2007voice}. To test the importance of transfer learning, we compared our system \vcsong~to \vcscratch~(trained on vocals from scratch), a model with same architecture but has not been pretrained with the knowledge from \vcvoice. Figs.~\ref{fig:GV1} and \ref{fig:MS1} present the comparison of GV and MS respectively between the two models compared to the ground truth. The root mean squared errors (RMSEs) between the models and the target are calculated and summarized in Table \ref{tab:rmse} showing that the RMSEs of \vcsong~with the target on the basis of GV and MS are smaller than those of \vcscratch. Consequently, transfer learning improves the average GV and MS by 35\% and 13\% respectively. % Finally, this proves that transfer learning improves the results of the conversion and is an integral part of the system.

\begin{figure*}[t]
\begin{center}
\begin{tabular}{cccc}
\includegraphics[width=0.32\textwidth]{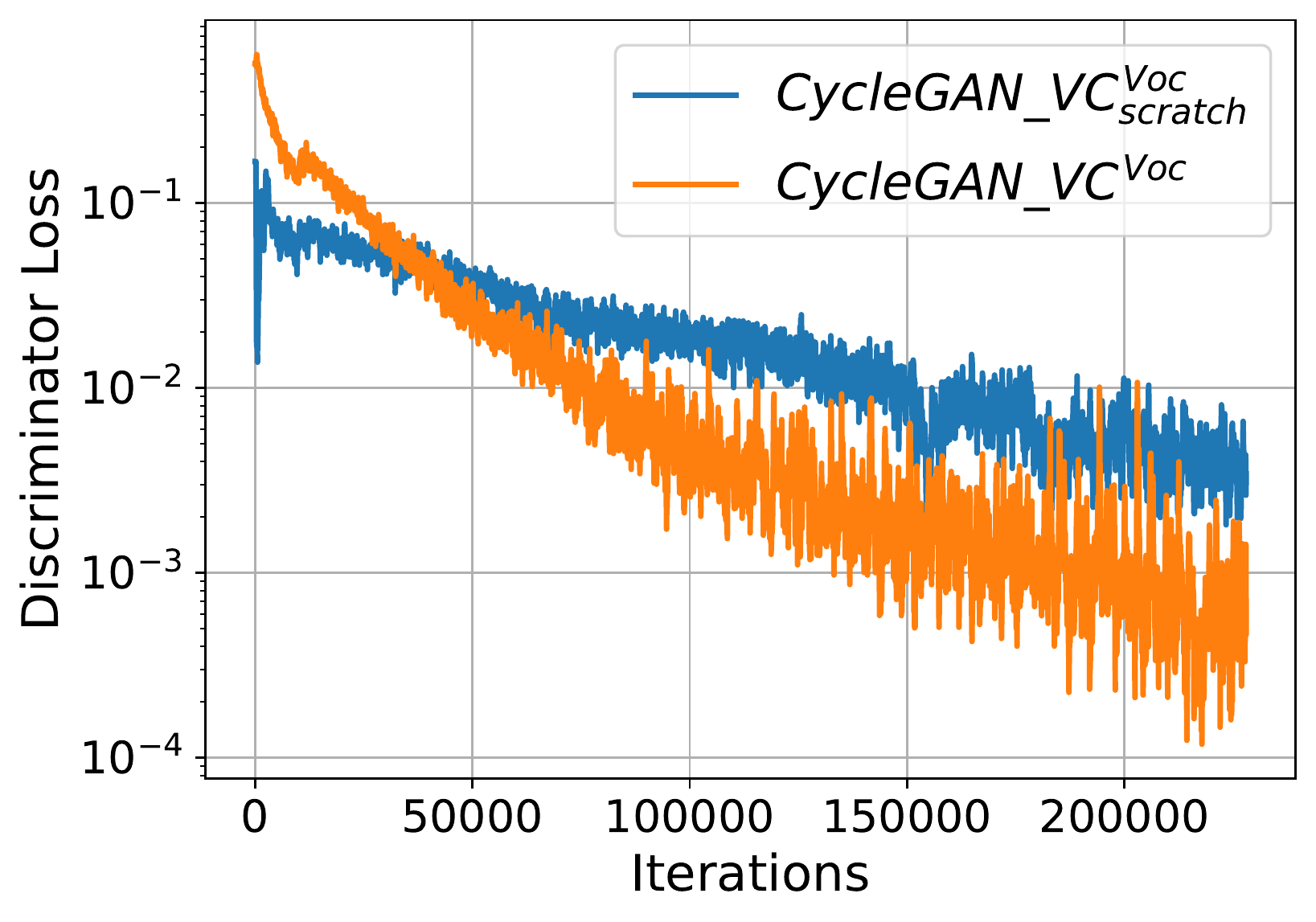}&
\includegraphics[width=0.32\textwidth]{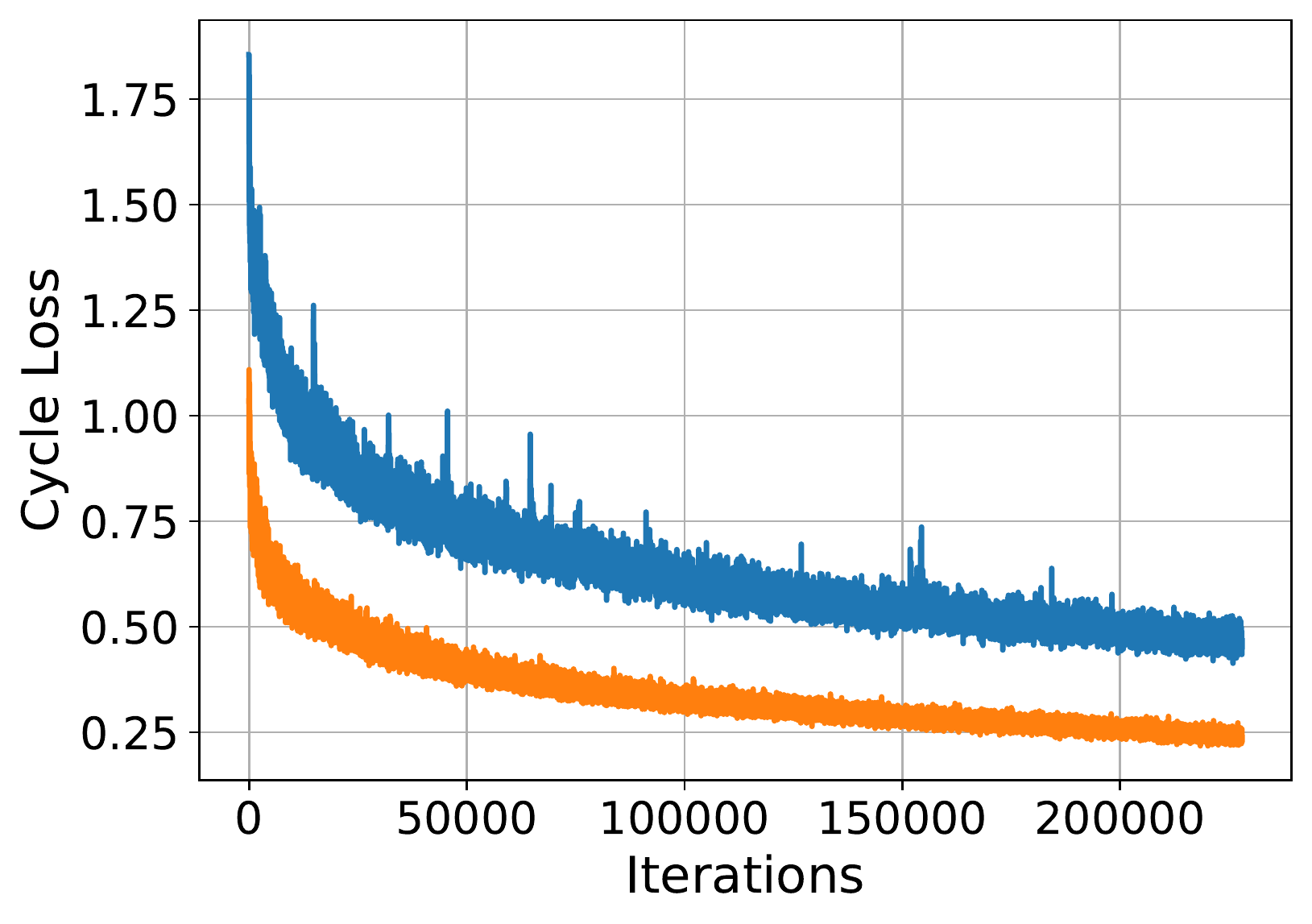}\\
\includegraphics[width=0.32\textwidth]{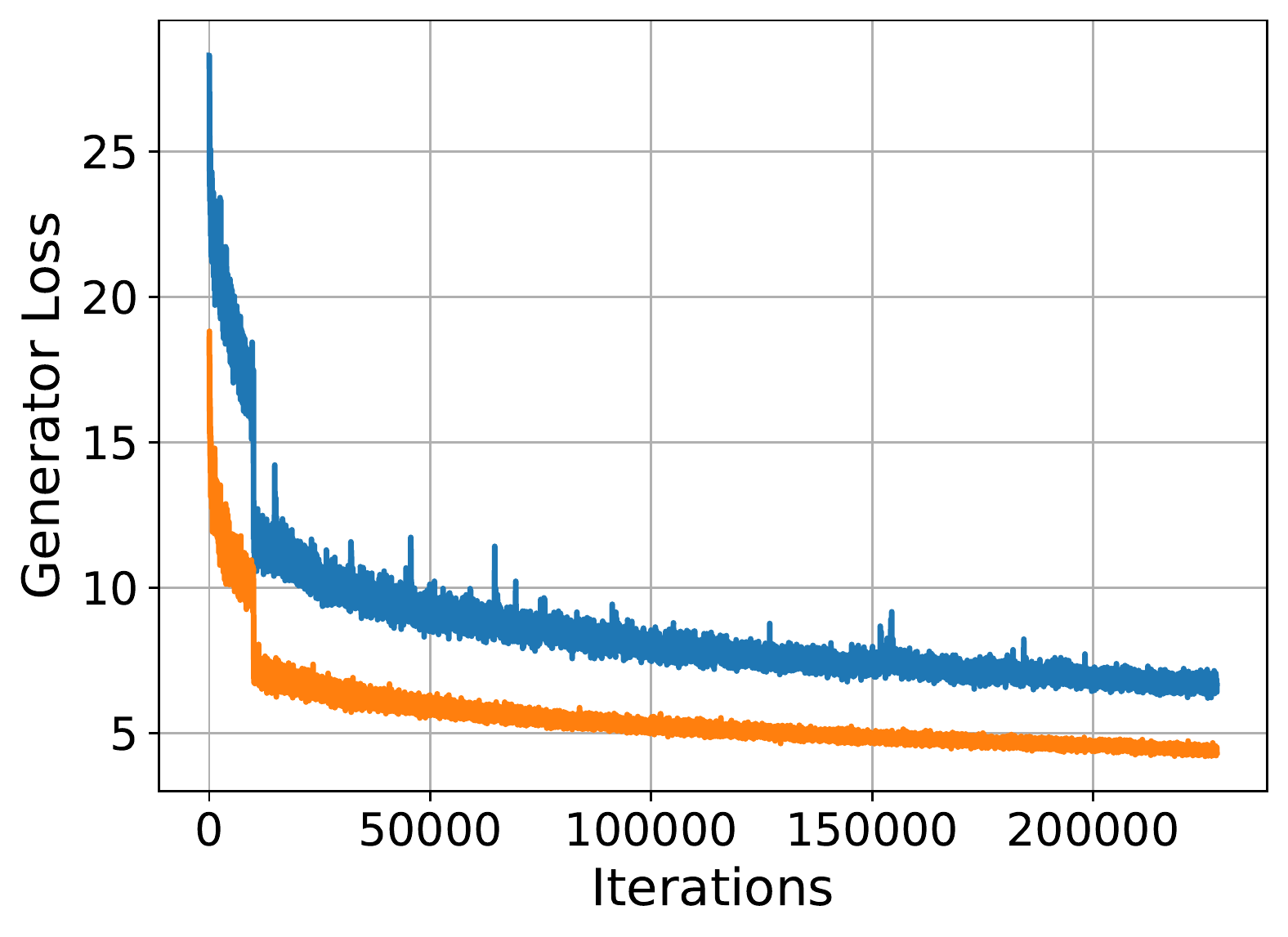}&
\includegraphics[width=0.32\textwidth]{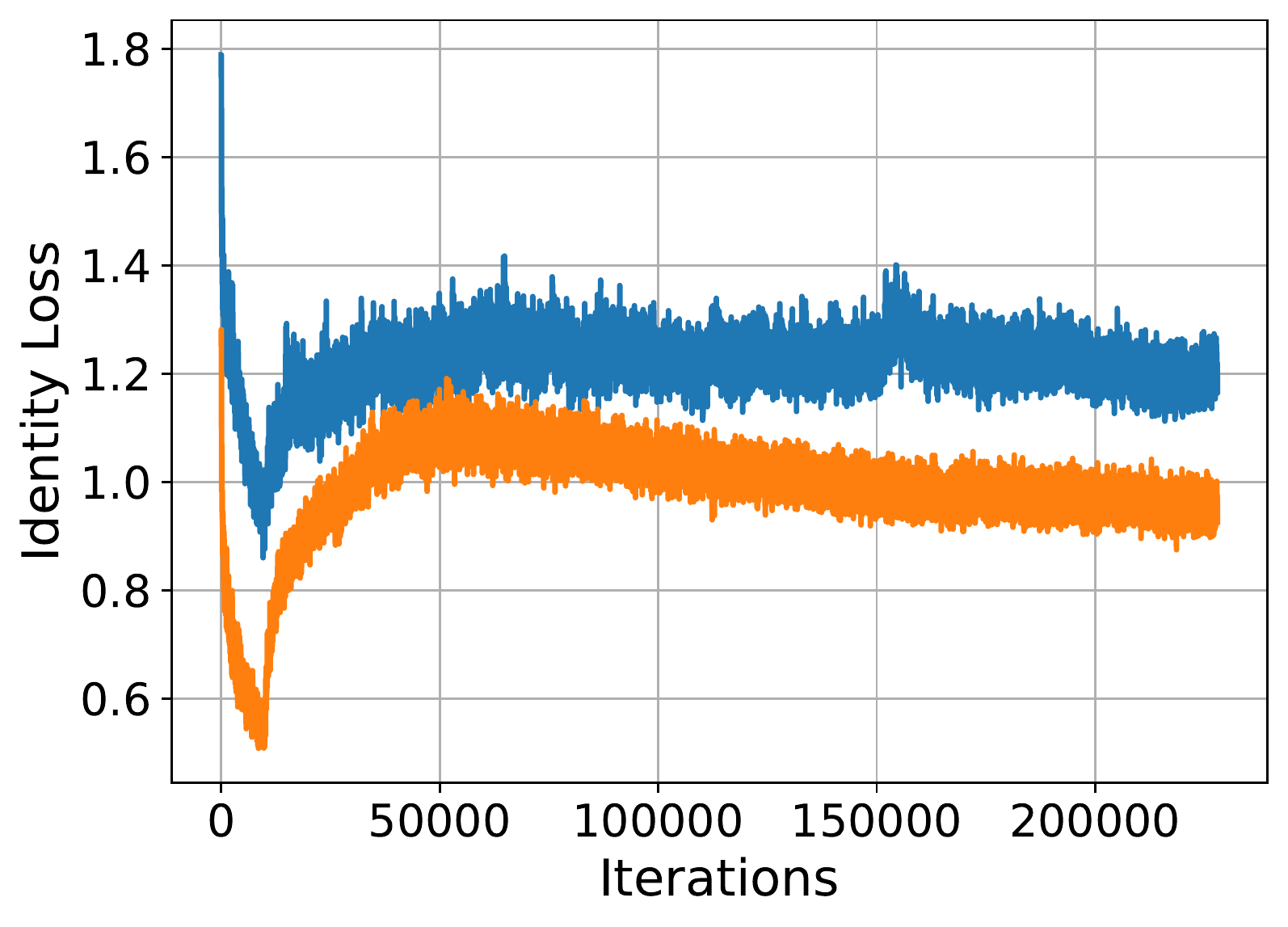}\cr
\end{tabular}
\end{center}
\vspace{-0.1cm}
\caption{Comparison of losses for \vcsong~and \vcscratch (best seen in color)}
\label{fig:losses}
\end{figure*}

To further validate the previous results, the change in \vcsong's losses throughout the training is compared to that of \vcscratch. It is worth mentioning that losses don't converge to zero, since GANs are used, which are nothing but a play on losses between the generator and the discriminator. Fig.~\ref{fig:losses} shows that the former outperformed the latter in terms of jump start performance and the final loss.% From these results, we conclude that using transfer learning in our pipeline enhanced our system.

%--------------------
%-------------------------

%--------------------

%------------------
%----------------------------------

%\begin{table}
%\centering
%\caption{Comparison of RMSE between target and converted GV and MS. Smaller values indicate resemblance to target.}
%\begin{tabular}{llllll}
%                 & \multicolumn{2}{l}{Segment 1} &  & \multicolumn{2}{l}{~Segment 2}  \\ 
%\hline
%Method           & ~GV           & ~MS                               &  & ~GV           & ~MS                                \\ 
%----------------------------------

%\hline
%\vcsong & \textbf{0.15} & \textbf{6.05}                     &  & \textbf{1.88} & \textbf{6.87}                      \\
%\vcscratch & 0.51          & 8.29                              &  & 2.40          & 7.95                              
%\end{tabular}\label{tab:score}
%\end{table}
%-------
%\begin{table}[htp]
%\centering
%\caption{Results of Performance Metrics for each Kernel}
%\begin{tabular}{|l|c| c|c|c|}
%\hline
%& \textbf{AC} & \textbf{RC} & \textbf{PR} & $\mathbf{F_1}$\\
%\hline
%RBF & \cellcolor[gray]{0.9} 87.91 & \cellcolor[gray]{0.9} 86.73 %& \cellcolor[gray]{0.9} 86.97 & \cellcolor[gray]{0.9} %86.82\\
%Poly& 61.7 & 59.49 & 67.67 & 55.46\\
%Linear  & 63.6 & 61.24 & 73.22 & 56.91\\
%\hline
%\end{tabular}
%\label{table:comparison}
%\end{table}

\begin{table}
\centering
\caption{Comparison of Average RMSE between target and converted instances in terms of GV and MS. Smaller values indicate resemblance to target.}
\begin{tabular}{lll}
Method & GV & MS  \\ 
\hline
\vcsong & \textbf{1.735} $\pm$ \textbf{1.076}  & \textbf{6.833} $\pm$ \textbf{0.373}\\
\vcscratch & 2.696 $\pm$ 2.386       & 7.922 $\pm$ 0.27 \\
\end{tabular}\label{tab:rmse}
\end{table}
%------------
\subsection{Subjective Evaluation}
We found it difficult to objectively test the effect of splitting the background music from vocals because the target ground truth song does not have the same background music nor the same pace. Hence, we subjectively evaluated the importance of splitting background music before being inputted to \vcsong, and we trained \vcnosplit~(trained on song, vocals and background music, using transfer learning), on full song segments without splitting, then compared it to \vcsong.  

To perform the subjective tests, we prepared a survey for song evaluation using five $10 sec$ song segments that were converted using two models: \vcsong~and \vcnosplit. The survey was filled by twenty test subjects of random gender, age, and musical background. The survey was conducted according to a ranking system on similarity basis ranging from $1$ (similar to original singer) to $5$ (similar to target speaker). Furthermore, naturalness was also included in the survey with a $1-5$ score ranging from not natural to very natural. 
The data from the survey was then analyzed using mean opinion score (MOS) test. The results in Table \ref{tab:MOS} show that the output of our system \vcsong~is closer to the target (reaching a 69\% similarity to the target) than \vcnosplit~(reaching only a 55\% similarity to the target) by 26\%. This was accompanied with an increase of 73\% in the degree of naturalness on average with our system \vcsong, and \vcnosplit~having 54\% and 31\% naturalness respectively. We demonstrated that our system \vcsong~has higher MOS than \vcnosplit. Particularly, we confirmed that data with background music has an adverse effect on the performance of the conversion model as expected, and adding a separation model to the pipeline had valuable implications on the output.
%--------------------
\begin{table}
\centering
\caption{Comparison of MOS of naturalness and similarity between target and converted. Larger values indicate resemblance to target and higher naturalness.}
\begin{tabular}{llll}
Method & Naturalness & & Similarity  \\ 
\hline
\vcsong & \textbf{2.68}                 &  & \textbf{3.46}\\
\vcnosplit & 1.55                       &  & 2.75
\end{tabular}\label{tab:MOS}
\end{table}

%---------------------
%Moreover, we also show the significance of training only on the vocals by comparing vcsong~to \vcnosplit. This was done by including in the survey the same five segments used before and converting them through \vcnosplit. 

\subsection{Limitations}
The analyzed results are also coupled with limitations that we will address in future work. The encountered limitations include the modest size of the dataset that had to be developed manually. That is, there is no ready dataset that includes two voices singing the same part with background music. Other limitations come from the drawbacks of using a subjective survey-based evaluation method, which may include dishonest answers, missing data, social desirability bias, unconscientious responses, and others.
\section{Conclusion}\label{sec:conc}
 In this paper, we presented our novel end to end framework,  \model~that successfully transformed songs to be performed by a target singer using an in-house collected dataset of non parallel songs. This was achieved by utilizing U-Net, Generative Adversarial Networks, and encoder-decoder architectures to first separate the songs from their background music, convert them, and then reconstruct the target song. The results were evaluated on the basis of the global variance and modulation spectrum of their corresponding Mel-spectrum coefficients which showed that transfer learning improves the performance of \model~by 35\%  in the global variance. The naturalness and similarity to the ground truth of the system output was evaluated with a subjective survey that shows the \model's output having 69\% similarity to the ground truth and 54\% naturalness. The encouraging results of our model SCM-GAN pave the way for an expansion into models that easily adapt to different target singers and languages through advanced forms of transfer learning. 
\section*{Acknowledgment}
We would like to thank the University Research Board of the American University of Beirut for funding this work, as well as Revotonix L.L.C for providing Amazon Web Services (AWS) credits for training the model. 

%\begin{thebibliography}{00}
%\bibitem{b1} G. Eason, B. Noble, and I. N. Sneddon, ``On certain integrals of Lipschitz-Hankel type involving products of Bessel functions,'' Phil. Trans. Roy. Soc. London, vol. A247, pp. 529--551, April 1955.
%\bibliography{refs}
%\end{thebibliography}
\bibliographystyle{IEEEtran}
\bibliography{refs}
\end{document}